\documentclass[lettersize,journal]{IEEEtran}
\usepackage{amsmath,amsfonts}
\usepackage{algorithm,algpseudocode}
\usepackage{array}
\usepackage[caption=false,font=footnotesize,labelfont=sf,textfont=sf]{subfig}
\usepackage{textcomp}
\usepackage{stfloats}
\usepackage{url}
\usepackage{verbatim}
\usepackage{graphicx}
\usepackage{cite}
\usepackage{tabularx}
\usepackage{makecell}
\usepackage{multirow}
\usepackage{cleveref}
\usepackage{booktabs}
\usepackage[dvipsnames]{xcolor}
\usepackage{colortbl}
\definecolor{Gray}{gray}{0.85}
\usepackage[inline]{enumitem}
\newenvironment{tableitemize}
{ \begin{minipage}[t]{\linewidth} \vspace{-10pt} \begin{itemize}[leftmargin=10pt] \vspace{5pt}}
{  \vspace{5pt} \end{itemize} \end{minipage}   } 

\begin{document}

\title{Digital Ethics in Federated Learning}

\author{Liangqi Yuan,~\IEEEmembership{Student Member,~IEEE},
        Ziran Wang,~\IEEEmembership{Member,~IEEE}, and
        Christopher G. Brinton,~\IEEEmembership{Senior Member,~IEEE}
\thanks{Manuscript received August 31, 2023.}
\thanks{L. Yuan, Z. Wang, and C. G. Brinton are with the College of Engineering, Purdue University, West Lafayette, IN 47907, USA (e-mails: liangqiy@purdue.edu; ryanwang11@hotmail.com, cgb@purdue.edu).}
}

\markboth{Journal of \LaTeX\ Class Files,~Vol.~14, No.~8, August~2021}%
{Shell \MakeLowercase{\textit{et al.}}: A Sample Article Using IEEEtran.cls for IEEE Journals}


\maketitle

\begin{abstract}
The Internet of Things (IoT) consistently generates vast amounts of data, sparking increasing concern over the protection of data privacy and the limitation of data misuse. Federated learning (FL) facilitates collaborative capabilities among multiple parties by sharing machine learning (ML) model parameters instead of raw user data, and it has recently gained significant attention for its potential in privacy preservation and learning efficiency enhancement. In this paper, we highlight the digital ethics concerns that arise when human-centric devices serve as clients in FL. More specifically, challenges of game dynamics, fairness, incentive, and continuity arise in FL due to differences in perspectives and objectives between clients and the server. We analyze these challenges and their solutions from the perspectives of both the client and the server, and through the viewpoints of centralized and decentralized FL. Finally, we explore the opportunities in FL for human-centric IoT as directions for future development.
\end{abstract}

\section{Introduction}
The Internet of Things (IoT) encompasses a phenomenon where physical devices, embedded with sensors, interact with their surroundings and engage in data exchange with other devices and systems via the Internet. These devices cover a vast range, from compact thermostats to large-scale industrial machinery, all interconnected within the IoT infrastructure. The human-centric IoT applications focus on devices within the IoT ecosystem that are designed to focus on human interaction or significantly influenced by human factors \cite{ystgaard2023review}, such as smartphones, wearable devices, vehicles, and healthcare appliances. These devices, through their diverse sensors, incessantly produce a wealth of highly sensitive data. For example, images taken by smartphones can contain global positioning systems (GPS) location information, smartwatches can detect a user's electrocardiogram (ECG), and vehicle navigation systems document a driver's routes. Certain companies might require customers to disclose such rich personal data to improve their machine learning (ML) models. The growing demand for data invariably raises concerns over potential privacy breaches and misuse of associated data, introducing pressing digital ethics issues in our interconnected digital era, such as privacy, security, and fairness.

Federated learning (FL) \cite{mcmahan2017communication} represents a decentralized learning paradigm, designed to facilitate multi-party collaboration while safeguarding user privacy. Its essence lies in sharing ML models rather than raw user data, achieving privacy preservation. Moreover, with the exponential growth of users and their data, the transmission and storage of vast amounts of raw data pose significant challenges for communication channels and server storage. FL can also be perceived as a form of knowledge distillation, distilling knowledge from raw data into model parameters to alleviate communication overhead. In line with the presence or absence of a server for coordination, management, and aggregation, FL can be categorized into two frameworks: centralized FL (CFL) and decentralized FL (DFL) \cite{yuan2023decentralized}. Initially proposed by Google researchers and deployed in the Google keyboard for cooperative learning in keyboard input recommendation models \cite{hard2018federated}, FL has found extensive applications in numerous sectors, such as healthcare, mobile services, and intelligent transportation systems, facilitating collaboration amongst widely distributed edge devices or institutions.

\begin{figure}[t]
\centering
\includegraphics[width=\linewidth]{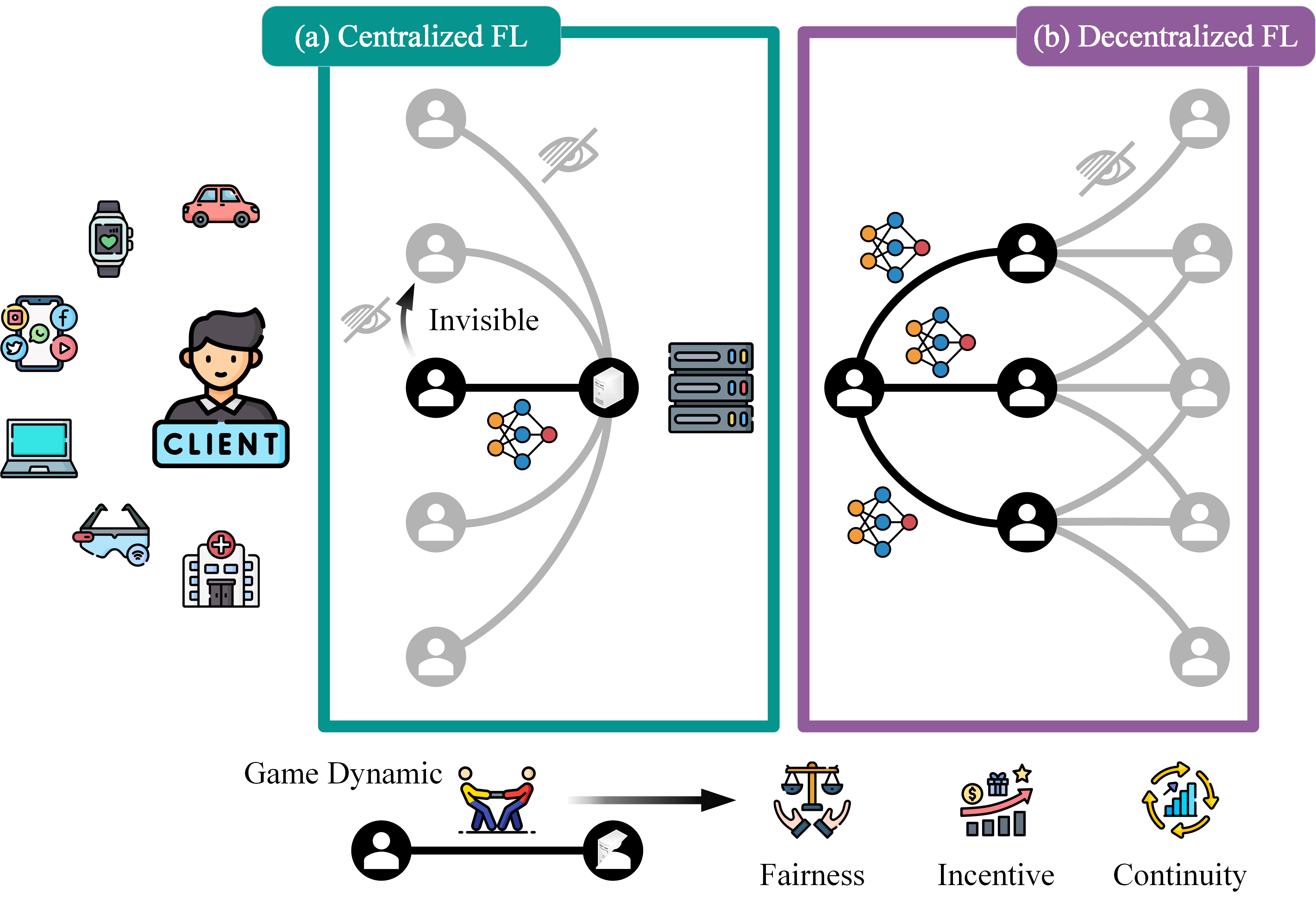}
\caption{The human-centric Internet of Things (IoT) applications within the (a) centralized federated learning (CFL) and (b) decentralized federated learning (DFL) frameworks.}
\label{Fig. Ethics} 
\end{figure}

FL presents a powerful approach for mitigating privacy concerns inherent in collaborative ML. However, digital ethical concerns extending beyond privacy are often overlooked \cite{chellapandi2023federated}, especially within the context of the human-centric IoT. Notably, most existing research inadequately addresses ethical considerations from the narrow perspective of the client side. For example, users of applications like Google Keyboard may remain oblivious to or unconcerned about the underlying FL algorithms. Their primary concern is that they are contributing their model but not receiving highly accurate personalized recommendations in return. This disparity in expectations can breed disappointment and potentially lead to a discontinuation of use. Consequently, human emotions may emerge as a vital factor in ensuring the continuity and longevity of FL frameworks in these contexts. 

\begin{figure*}[t]
\centering
\includegraphics[width=\linewidth]{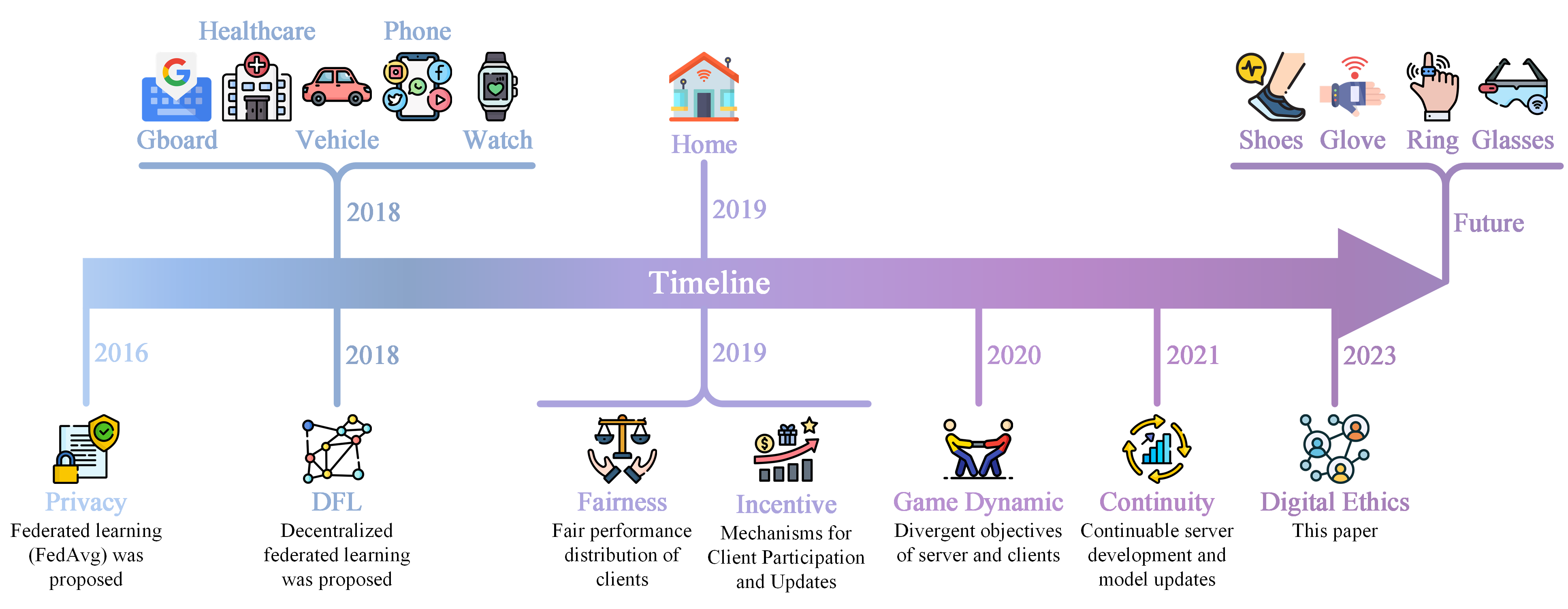}
\caption{Lifeline of digital ethics and the human-centric Internet of Things (IoT) applications in federated learning (FL).}
\label{Fig. Roadmap} 
\end{figure*}

In this paper, we present a discourse on the digital ethical issues arising within both CFL and DFL deployments in human-centric IoT applications, as depicted in Fig. \ref{Fig. Ethics}. We illustrate the FL lifeline in Fig. \ref{Fig. Roadmap}, encompassing two trajectories, namely human-centric IoT and the digital ethics of FL. Apart from user privacy, people are generally concerned about fairness, interpretability, accountability, transparency, and other aspects. Additionally, issues related to user management, incentives, penalties, continuity, and compatibility with new users are important considerations in FL systems. In addition to pursuing higher performance and convergence in ML and optimizing communication networks, there is also a growing interest in the social, psychological, and economic aspects of FL, among others.

The organization of this paper is as follows: First, we provide an in-depth examination of the definitions and perspectives of clients and the server, as well as the underlying reasons for the game dynamic relationship that arises between the CFL and DFL frameworks (Sec. \ref{Sec. Variance of Perspectives}). The discrepancies, limitations, and information asymmetry between clients and the server, especially the fundamental difference in their objectives, inevitably give rise to a game dynamic (Sec. \ref{Sec. Game Dynamics}). Subsequently, the resultant trust issues emerging from divergent objectives appear specifically as client skepticism towards the fairness of the FL framework (Sec. \ref{Sec. Fairness}). Notably, the difference in perspective also leads to varied definitions of fairness between clients and the server. Adjacent to the issue of fairness is the problem of incentive mechanisms for clients (Sec. \ref{Sec. Incentives}). Beyond the extensively researched server-led incentive mechanisms, we discuss the potential for a reputation system, established by the client community, to become one of the primary mechanisms in DFL. Alongside fairness and incentives, we also touch upon the continuity of FL's development and updates (Sec. \ref{Sec. Continuity}). Based on these four properties (i.e., game dynamics, fairness, incentives, and continuity), we proceed to discuss opportunities to foster the continuous, active, and positive development of FL (Sec. \ref{Sec. Opportunities}). Finally, we draw conclusions from this paper (Sec. \ref{Sec. Conclusion}).

\section{Variance of Perspectives}
\label{Sec. Variance of Perspectives}

Within FL, different roles possess distinct perspectives and varied levels of knowledge. The core of the game dynamics in FL stems from the differences in clients' contributions (e.g., the volume of raw data), the learning process, and the information asymmetry among participants.

\subsection{Omniscient (Authors' and Readers') Perspective}

Currently, a significant portion of research papers on FL tends to overlook the information asymmetry between clients and servers. They often adopt an idealized perspective, optimizing FL based on the assumption of complete and perfect knowledge. These design methods, founded on the notion of omniscient information, fail to address the practical challenges that arise from limited information exchange, client data heterogeneity, and potential trust issues. Recognizing and considering the scenarios of information asymmetry are crucial for developing effective FL systems.

\subsection{Server's Perspective}

In CFL, the role of the server is to receive model parameters from all clients, aggregate them, and then redistribute the aggregated model. The server, however, remains oblivious to how clients collect data, train models, and handle the post-processing of models. Some CFL frameworks make presumptions that the server is privy to more extensive metadata from clients. For example, in the case of FedAvg, each client not only sends their model parameters but also transmits the volume of their local raw data. This additional metadata allows the server to perform weighted averaging. Therefore, in CFL, the resources or perspectives available to the server can be summarized as previously aggregated models, current and past models from clients, and other metadata that clients are asked to send, such as volume of raw data, performance on local test sets, losses, training epochs, etc.

\subsection{Clients' Perspective}

Considering the perspective of the clients, we discuss the contexts of both CFL and DFL frameworks, as illustrated in Fig. \ref{Fig. Ethics}.

In CFL, clients are oblivious to each other's information, such as the number of clients, the volume of raw data each client holds, the learning process, and the model performance of each. In specific FL scenarios, for example, when healthcare institutions act as clients for FL, the components such as optimizers, loss functions, learning rates, and training epochs differ from client to client. Additionally, clients lack knowledge about the server-side details, like the aggregation method employed by the server. Hence, in a CFL framework, the only available information for each client is the aggregated model received from the server.

In DFL, clients directly share models without the coordination of a server. For example, in a fully connected network topology, every client within the DFL framework needs to transmit their own model parameters to all other clients, and reciprocally receive models from them. As a result, a certain framework, such as network topology, communication direction, frequency, and so forth, needs to be agreed upon among clients. They also need to be cognizant of certain information about other clients, like their addresses and ports. Additionally, some extra metadata, such as the volume of raw data, number of clients, model versions, etc., might also be transmitted as per the requirement. Therefore, in DFL, for the system to function correctly, clients need to establish a communication protocol among themselves and are required to directly disclose their local information to other clients.

\begin{table*}[t]
\caption{Definition of Digital Ethics for Client and Server in CFL and DFL}
\label{Table Sociology of FL}
\centering
\begin{tabularx}{\linewidth}{|l|p{70pt}|p{65pt}|X|p{95pt}|p{85pt}|}
\toprule
~ & \textbf{Perspective} & \textbf{Game Dynamics (Objective)} & \textbf{Fairness} & \textbf{Incentives} & \textbf{Continuity}  \\
\midrule
\multicolumn{6}{|c|}{\textbf{Centralized Federated Learning}} \\
\midrule
\textbf{Omniscient} & Everything & Generalized and personalized model & Fairness-aware strategy & Incentive mechanism design & Server Management Enhancement \\
\midrule
\textbf{Server} & Models and information from the clients & Generalized model & Overall accuracy is highest & Revenue, market share, available generalized models & Management of clients and models \\
\midrule
\textbf{Client} & Model from the server & Personalized model & Local accuracy is highest & Rewards, punishments, and model updates & Compliance with server management \\
\midrule
\multicolumn{6}{|c|}{\textbf{Decentralized Federated Learning}} \\
\midrule
\textbf{Omniscient} & Everything & Generalized and personalized model & Fairness-aware strategy & Incentive mechanism design & Encouraging spontaneous management by clients \\
\midrule
\textbf{Server} & \multicolumn{5}{c|}{N/A} \\
\midrule
\textbf{Client} & Model from other clients & Personalized model & Local accuracy is highest & Exposure, reputation, and model updates & Identify, accuse, and report malicious clients \\
\bottomrule
\end{tabularx}
\end{table*}

\section{Game Dynamics}
\label{Sec. Game Dynamics}

\subsection{Why Game Dynamic Emerge?}

Compared with distributed learning that assigns tasks to nodes or miners, FL inherently emphasizes more on the data-generating clients. Governed by these data-holding clients, and propelled by self-interest and greed, the inclination towards selfish behavior and a lack of trust in others could surface. This drives the dynamics of interaction among clients in a game dynamics context, where each seeks to maximize personal gains.

This dynamic is primarily attributed to significant data heterogeneity among clients, where the server-aggregated model may not exhibit exceptional performance on all clients. Firstly, inter-group heterogeneity exists among clients. For example, professionals such as professors, doctors, and lawyers using Google Keyboard would require highly tailored recommendations due to their distinctive fields of expertise. Secondly, intra-group heterogeneity exists within each group of users, wherein each user's academic discipline, level of knowledge, years of expertise, and other factors vary. Lastly, system heterogeneity among clients arises from variations in IoT devices, which includes disparities among sensor and instrument manufacturers, differences in software versions of devices, and varying user operations.

\subsection{Game Dynamics between Client and Server}

In CFL, clients and the server share similar yet fundamentally different objectives: clients aim to achieve the best-performing personalized model on their local dataset, while the server seeks to achieve the best average-performing general model across all clients. While this setup can be mutually beneficial, a game dynamics relationship emerges between the clients and the server due to the trade-off between personalized performance and generalization. 

Personalized FL represents a potential solution to mitigating the game dynamics between clients and the server, as it seeks to satisfy the objectives of both parties \cite{tan2022towards}. There are two main strategies in this context: client compromise and server compromise. In the case of client compromise, a simplistic implementation would involve the client performing additional gradient descent upon receiving the generalized global model (i.e., meta-learning), thus achieving personalized expansion \cite{yuan2023federated}. Conversely, in server compromise, a common practice is clustered FL. In this scenario, the server can create multiple aggregated global models based on the nature of the clients or clustering of client models, with even the potential for multiple servers to partition different regions for aggregation (i.e., hierarchical FL). Regardless of whether it is client compromise or server compromise, these strategies both entail additional overheads, such as computation, communication, and storage.

\subsection{Game Dynamics among Clients}

In DFL, despite the symmetric roles of clients in communication and knowledge propagation, competition can still emerge due to data heterogeneity and system heterogeneity. They all share the same but conflicting objective of striving for the best model performance on their respective local data sets. However, given the data heterogeneity among clients, it is more common that the models from other clients perform poorly on their local dataset \cite{yuan2023peer}.

Analogous to the two compromise strategies in CFL, DFL also incorporates similar personalized methods to enhance the performance of aggregated models on clients' local data sets. Apart from model post-processing methods such as meta-learning, DFL can reduce data heterogeneity among clients within a cluster by establishing different topological structures, mimicking the clustered FL. In particular, in real-world DFL scenarios, clients might prefer freely forming their clusters and establishing DFL network topologies among similar and familiar populations, such as city clusters and suburb clusters determined by geographical locations. These more flexible and customizable network topologies, although more challenging to establish initially, also confer more personalized and trustworthy DFL with communication cost advantages.

\begin{table*}[t]
\caption{State-of-the-art technologies, strategies, and mechanisms}
\label{Table State-of-the-art}
\centering
\begin{tabularx}{\linewidth}{|p{60pt}|X|p{250pt}|}
\toprule
\textbf{Issue} & \textbf{Definition} & \textbf{Technologies, strategies, and mechanisms} \\
\midrule
\textbf{Game Dynamics (Objective)} & Divergent objectives amongst clients and server & \begin{tableitemize} \item Personalized FL \begin{itemize} \item Server compromise (e.g., clustering, knowledge distillation) \item Client compromise (e.g., meta-learning, data augmentation) \end{itemize} \end{tableitemize} \\
\midrule
\textbf{Fairness} & Contribution and performance distribution among clients & \begin{tableitemize} \item Averaging (i.e., arithmetic mean) \item Weighted averaging (e.g., FedAvg is based on sample volume) \item Post-processing (e.g., personalized CFL) \end{tableitemize} \\
\midrule
\textbf{Incentives} & Clients contribute honestly, actively, and positively & \begin{tableitemize} \item Feedback (e.g., aggregated model) \item Reward (e.g., sponsorship, subscription) \item Reputation (e.g., like, follow, share) \end{tableitemize} \\
\midrule
\textbf{Continuity} & FL maintains its operations and efficiency to prolong its lifecycle & \begin{tableitemize} \item Enlistment of new clients \item Rapid and efficient model iteration \item Low computational, communication, and storage overheads \end{tableitemize}\\
\bottomrule
\end{tabularx}
\end{table*}

\section{Fairness}
\label{Sec. Fairness}

\subsection{How to Define Fairness?}
A fairer system would enhance client trust, incentivize client contributions, reduce the potential for free-riding behaviors, attract more new client engagement, and bolster the long-term continuity of the system, among other benefits. Fairness has always been a central theme in human cooperation, and FL is no exception. Particularly in CFL, the method of aggregation has spurred discussions concerning fairness. The question of fairness in FL is indeed become a focal point of discourse. It involves considering whether the FL framework should prioritize the majority of users and clients with a larger number of samples, or whether it should also take into account clients with fewer samples that may have lower representativeness \cite{li2019fair}. Furthermore, the perspectives of both clients and the server may also sway their interpretations of fairness. Clients may not have full visibility into the server's aggregation algorithm, nor comprehend the performance of the aggregated model across different clients.

\subsection{Fairness of Server}

From the perspective of the server, its objective is to pursue a generalized model that maximizes the overall average performance of all clients. Driven by this objective and the pursuit of generalization in FL, the server's concept of fairness often tends to favor clients with a greater influence or voice. In the classic FedAvg algorithm, the server conducts a weighted averaging aggregation based on the sample number of each client. This approach seems fair because the sample size can extent reflect the performance and credibility of the model to some extent, and can be seen as a reward for clients with more samples, as the aggregated model is more likely to bias towards them. However, for those underrepresented clients, the performance of the aggregated model might be unsatisfactory. Furthermore, the dominance of large sample clients could lead to low sample diversity and cause the aggregated model to lose its generalization capability. On the contrary, an FL framework involving non-weighted averaging during aggregation might demotivate clients with large sample sizes, subsequently diminishing system performance and continuity. Hence, from the server's standpoint, the conception of fairness remains a topic open to debate.

\subsection{Fairness of Client}

Conversely, from the client's perspective, their interpretation of fairness tends to be simpler. This is primarily due to the likelihood that they are either unaware of or unconcerned with the server's aggregation algorithm and the performance of the aggregated model on other clients. Hence, within the FL framework, clients would consider the system fair from a standpoint of individual fairness, provided the model maintains acceptable performance locally. A noteworthy example is Google Keyboard, where users contribute local model parameters in their usage, and in return, benefit from personalized recommendations. Interestingly, these recommendations from Google Keyboard are not necessarily completely accurate. As long as the output is within the user's range of acceptability, the application can maintain its advantage relative to non-personalized keyboard applications. Of course, it is crucial to note that the level of acceptable performance may vary according to individual clients' requirements and should not be generalized. When using Google Keyboard, users are often oblivious to or indifferent towards the server's aggregation algorithm and have minimal or no knowledge of the model's performance among other users. Users are likely to be self-interested, prioritizing their user experience without considering any factors related to others, i.e., a non-cooperative game scenario.

\section{Incentives}
\label{Sec. Incentives}

\subsection{Incentives Driven by Server}

As the owner, leader, and manager of FL frameworks, the server typically aspires for its framework to undergo large-scale, active, positive, and continuable development. Thus, how the server employs incentive strategies to encourage clients to report their models, metadata, contributions, and even flaws in a rational, honest, and proactive manner remains an unresolved issue. The right to use the aggregated model itself serves as a form of reward (passive incentive), and current incentive strategies also contemplate offering additional rewards disseminated by the server to motivate clients (active incentives). The client contributions in these incentive strategies can follow economic principles, such as game theory, auction, contract, matching theory, and so forth \cite{tu2022incentive}. The Stackelberg game, in particular, has garnered considerable attention due to its alignment with the behaviors of the server and clients in a CFL setting. Besides these active incentives, punitive incentives may also serve as a potential strategy. For example, the right to use the aggregated model could be revoked if a client's contribution does not meet expectations.

\subsection{Incentives Driven by Client Community}

Within the context of DFL, the absence of server coordination and the customizability of diverse network topologies render the incentive problem more variable and challenging \cite{chellapandi2023survey}. On the one hand, there is no server to generate and distribute rewards, while on the other hand, calculating client contributions is especially difficult due to mutual distrust among clients. Therefore, certain passive incentive strategies may become more effective and prevalent than active incentives. On one side, clients can acquire the right to use the models by participating in the DFL community. Simultaneously, due to the factors of information asymmetry and mutual invisibility of information among clients, they are unaware of the size of each other's contributions, such as the volume of raw data, training epochs, optimization results, etc. Consequently, they might be more inclined to share models imbued with local knowledge in exchange for other clients' model updates. The motivation here is to garner as many resources as possible from the client community, albeit at the expense of disclosing local resources. We can draw inspiration from altruistic contribution behaviors observed in human societies, such as open-sourcing on Github, answering questions on Stack Overflow, voluntarily performing peer reviews, etc. While free-riding attacks (where some users garner knowledge from others without contributing themselves) are inevitable, the influence of reputation and prestige can nonetheless maintain a virtuous cycle within the community \cite{fehr1999theory}.

\section{Continuity}
\label{Sec. Continuity}

Continuity is a critical feature for the survival, revenue generation, and expansion of any application, system, or framework. In terms of FL, continuity signifies the pause, elimination, and reactivation of inactive clients, the continual, active, and voluntary updates from current clients, and the willingness, eligibility, and data diversity of a large number of prospective clients.

\subsection{Continuable Development of Server}

From the server's perspective, continuable development necessitates addressing and responding to the needs of these three classes of clients  - inactive, current, and potential - while also considering the maintenance of different versions of the model to prevent catastrophic forgetting. More specifically, due to the continuous generation of new data by clients in the real world, particularly IoT devices, the local model updates of clients are typically based on the latest data. Although new models are evidently more compelling due to factors such as scenario updates, user utilization, and concept drift, old versions of the models do not entirely lose their contributions. A potential example could be an application using IoT devices, such as a smartwatch that monitors user's ECG patterns. The ECG readings of users are likely to differ between weekdays and weekends, thus the models derived from weekend data might warrant individual storage. In practice for FL, while the server is aggregating the current versions of local models, it also incorporates previous versions with appropriate weighting. Furthermore, clients are granted the ability to trace back and retrieve prior versions of the model at any time. This feature serves as a safeguard against potential instability in client performance due to model updates.

\subsection{Continuable Update of Client}

In the context of CFL, clients strive for long-term stability, rapid iterations, and efficient updates of the aggregated models from the server. Thus, they may work hard to deploy server-updated models at the earliest opportunity to achieve enhanced performance and user experience. Beyond their expectations from the server, under rational circumstances, clients might also attempt to report their model parameters to the server as rapidly, thoroughly, and accurately as possible, to ensure their models are significantly considered during the server's aggregation process. This is because the server cannot indefinitely wait for all clients to upload their models. Therefore, in a rational state, the behavior of client updates is balanced between the long-term nature of data collection and the rapidity of model updates.

In the scenario of DFL, clients within the community may voluntarily identify, denounce, and report malicious clients performing adversarial attacks (e.g., model poisoning) in order to protect the community, given that this relates to their own interests. This is because the incorporation of models from these malicious clients into the FL process could potentially harm their interests. Clients might also proactively share their models with other clients, establishing a good reputation, so that other clients will be inclined to promptly share their model updates in return. One potential concern is that the DFL client population may exhibit exclusionary tendencies. Specifically, the mistrust towards new clients and the uncertainty brought about by their models, especially within smaller communities, can be quite pronounced. This may further hinder the continuity and growth of such small-scale communities.

\section{Opportunities}
\label{Sec. Opportunities}

\subsection{Interplay of Game Dynamics, Fairness, Incentives, and Continuity}

The issues of game dynamics, fairness, incentive mechanisms, and continuity in FL are interrelated and mutually impactful. For example, if an FL framework could perfectly achieve the objectives of all clients, such as Google Keyboard ideally meeting user expectations, users would naturally diminish their concerns about fairness. A fairness-aware strategy can also be considered as an incentive mechanism where clients contributing more are rewarded proportionally. Taking FedAvg as an example, clients might make great efforts to contribute as much local data as possible to the model training, to gain a more significant voice during the server's model aggregation process. Therefore, fairness-aware strategies of weighted aggregation indirectly incentivize clients to make more contributions. Concurrently, this enhances the continuity of the FL framework, as each client will make an effort to collect data, train models, and participate in FL updates promptly to gain rewards. Under such continuable conditions, the game dynamics within the FL framework are also mitigated, as each client generates a steady stream of data resources, enabling the training of more robust models. Therefore, for the issues of game dynamics, fairness, incentive mechanisms, and continuity, both parallel multi-solution approaches and single-solution breakthroughs are viable options.


\subsection{Integration with Sociology and Ethology}

FL essentially represents a form of knowledge propagation, a method that is already widespread, diverse, and matured within both human societies and animal behaviors \cite{eibl2017human}. For example, the instructive paradigm between a teacher and students can offer insights to CFL, resonating with the architecture of a large model within the server and smaller models among clients utilized in federated knowledge distillation \cite{wu2022communication}. Intriguingly, a similar hierarchical structure is observed in the field of ethology, particularly within ant colonies or bee hives. Here, directives (models) from the queen ant or queen bee (server) are disseminated to the worker ants or bees (clients), offering a clear instance of role distribution.

DFL is increasingly becoming a focus for researchers, due to its capacity to circumvent limitations imposed by server dependency, and also its reflection of more prevalent modes of knowledge dissemination among clients within human societies and ethology. For example, in the context of conferences, speakers (clients) present their research findings (models) to all attendees (other clients), which can be viewed as a manifestation of fully connected DFL. In group collaborations, each team member (client) contributes a part towards a common goal (model), mirroring the concept of split DFL. Interestingly, similar decentralized patterns of knowledge dissemination are observable in animal behavior. For example, within a school of fish, individual fish (clients) only communicate with their neighbors (gossip protocol), but when danger arises, the alert signal (model) spreads across the entire school (other clients), promoting swift collective evasion \cite{sayed2013diffusion}.

Therefore, incorporating insights from sociology and ethology can effectively enhance FL organizational structures that are centered on IoT users, better aligning with the psychological expectations of users as clients.

\subsection{Deployment Optimization in Federated Learning}

Current research mainly centers on the optimization of training and communication within FL, largely overlooking the strategy and timing for deploying the base model in FL on client devices. Specifically, in classical algorithms such as FedAvg, the fundamental operational cycle entails download $\rightarrow$ train $\rightarrow$ upload $\rightarrow$ download $\rightarrow$ deploy. In contrast, personalized algorithms, such as meta-learning, follow a cycle of download $\rightarrow$ train $\rightarrow$ upload $\rightarrow$ download $\rightarrow$ train $\rightarrow$ deploy. Therefore, it's evident that the deployment sequence within the communication and training processes significantly affects the performance of the model. With this in mind, we propose considering two distinct deployment sequences, facilitating the deployment of either generalized or personalized models, contingent on the specific use case:
\begin{enumerate}[label=(\roman*)]
\item Deploy post-download for a generalized model: The model deployed is the aggregated one, offering wider generalization capabilities. However, it may not necessarily deliver optimal performance on local datasets.
\item Deploy post-training for a personalized model: The model deployed is the one locally trained on the aggregated model, offering a higher degree of personalization and subsequently, enhancing confidence in the model's performance.
\end{enumerate}
Beyond the influence of the order of deployment on performance within the FL process, in the real world, due to the sequential and time-sensitive nature of data collection from IoT devices, excessive waiting for responses from the server or other clients may degrade model performance. Therefore, deployment optimization in FL, as an issue rooted in real-world applications, builds upon the foundational capacities of training and communication to further enhance FL's performance, credibility, and operational efficiency.

\section{Conclusion}
\label{Sec. Conclusion}

In this paper, we explore and discuss FL in the context of human-centric IoT applications, with a particular emphasis on the advancements made by FL algorithms in addressing human privacy concerns, as well as other digital ethical dilemmas. We take into account perspectives from three distinct roles: the omniscient, clients, and the server, with a detailed analysis of both the CFL and DFL frameworks. Each of these roles, characterized by varying objectives and information asymmetries, raises game dynamics and trust crises, which in turn incite debates around fairness, incentive, and continuity. This paper aims to highlight the prevalent disregard for human digital ethics in the current FL paradigm and to inspire the future design of FL frameworks from sociological, psychological, and economic perspectives.

\bibliographystyle{IEEEtran}
\small\bibliography{reference}

\begin{thebibliography}{10}
\providecommand{\url}[1]{#1}
\csname url@samestyle\endcsname
\providecommand{\newblock}{\relax}
\providecommand{\bibinfo}[2]{#2}
\providecommand{\BIBentrySTDinterwordspacing}{\spaceskip=0pt\relax}
\providecommand{\BIBentryALTinterwordstretchfactor}{4}
\providecommand{\BIBentryALTinterwordspacing}{\spaceskip=\fontdimen2\font plus
\BIBentryALTinterwordstretchfactor\fontdimen3\font minus \fontdimen4\font\relax}
\providecommand{\BIBforeignlanguage}[2]{{%
\expandafter\ifx\csname l@#1\endcsname\relax
\typeout{** WARNING: IEEEtran.bst: No hyphenation pattern has been}%
\typeout{** loaded for the language `#1'. Using the pattern for}%
\typeout{** the default language instead.}%
\else
\language=\csname l@#1\endcsname
\fi
#2}}
\providecommand{\BIBdecl}{\relax}
\BIBdecl

\bibitem{ystgaard2023review}
K.~F. Ystgaard, L.~Atzori, D.~Palma, P.~E. Heegaard, L.~E. Bertheussen, M.~R. Jensen, and K.~De~Moor, ``Review of the theory, principles, and design requirements of human-centric internet of things (iot),'' \emph{J. Ambient Intell. Humaniz. Comput.}, vol.~14, no.~3, pp. 2827--2859, Mar. 2023.

\bibitem{mcmahan2017communication}
B.~McMahan, E.~Moore, D.~Ramage, S.~Hampson, and B.~A. y~Arcas, ``Communication-efficient learning of deep networks from decentralized data,'' in \emph{Artificial intelligence and statistics}.\hskip 1em plus 0.5em minus 0.4em\relax PMLR, 2017, pp. 1273--1282.

\bibitem{yuan2023decentralized}
L.~Yuan, L.~Sun, P.~S. Yu, and Z.~Wang, ``Decentralized federated learning: A survey and perspective,'' \emph{arXiv preprint arXiv:2306.01603}, Jun. 2023.

\bibitem{hard2018federated}
A.~Hard, K.~Rao, R.~Mathews, S.~Ramaswamy, F.~Beaufays, S.~Augenstein, H.~Eichner, C.~Kiddon, and D.~Ramage, ``Federated learning for mobile keyboard prediction,'' \emph{arXiv preprint arXiv:1811.03604}, Nov. 2018.

\bibitem{chellapandi2023federated}
V.~P. Chellapandi, L.~Yuan, C.~G. Brinton, S.~H. Zak, and Z.~Wang, ``Federated learning for connected and automated vehicles: A survey of existing approaches and challenges,'' \emph{arXiv preprint arXiv:2308.10407}, 2023.

\bibitem{tan2022towards}
A.~Z. Tan, H.~Yu, L.~Cui, and Q.~Yang, ``Towards personalized federated learning,'' \emph{IEEE Trans. Neural Netw. Learn. Syst.}, Mar. 2022.

\bibitem{yuan2023federated}
L.~Yuan, L.~Su, and Z.~Wang, ``Federated transfer-ordered-personalized learning for driver monitoring application,'' \emph{arXiv preprint arXiv:2301.04829}, Jan. 2023.

\bibitem{yuan2023peer}
L.~Yuan, Y.~Ma, L.~Su, and Z.~Wang, ``Peer-to-peer federated continual learning for naturalistic driving action recognition,'' in \emph{Proceedings of the IEEE/CVF Conference on Computer Vision and Pattern Recognition}, 2023, pp. 5249--5258.

\bibitem{li2019fair}
T.~Li, M.~Sanjabi, A.~Beirami, and V.~Smith, ``Fair resource allocation in federated learning,'' \emph{arXiv preprint arXiv:1905.10497}, May 2019.

\bibitem{tu2022incentive}
X.~Tu, K.~Zhu, N.~C. Luong, D.~Niyato, Y.~Zhang, and J.~Li, ``Incentive mechanisms for federated learning: From economic and game theoretic perspective,'' \emph{IEEE Trans. Cogn. Commun. Netw.}, May 2022.

\bibitem{chellapandi2023survey}
V.~P. Chellapandi, L.~Yuan, S.~H. Zak, and Z.~Wang, ``A survey of federated learning for connected and automated vehicles,'' \emph{arXiv preprint arXiv:2303.10677}, Mar. 2023.

\bibitem{fehr1999theory}
E.~Fehr and K.~M. Schmidt, ``A theory of fairness, competition, and cooperation,'' \emph{Q. J. Econ.}, vol. 114, no.~3, pp. 817--868, Aug. 1999.

\bibitem{eibl2017human}
I.~Eibl-Eibesfeldt, \emph{Human ethology}.\hskip 1em plus 0.5em minus 0.4em\relax Routledge, 2017.

\bibitem{wu2022communication}
C.~Wu, F.~Wu, L.~Lyu, Y.~Huang, and X.~Xie, ``Communication-efficient federated learning via knowledge distillation,'' \emph{Nat. Commun.}, vol.~13, no.~1, p. 2032, Apr. 2022.

\bibitem{sayed2013diffusion}
A.~H. Sayed, S.-Y. Tu, J.~Chen, X.~Zhao, and Z.~J. Towfic, ``Diffusion strategies for adaptation and learning over networks: an examination of distributed strategies and network behavior,'' \emph{IEEE Signal Process. Mag.}, vol.~30, no.~3, pp. 155--171, Apr. 2013.

\end{thebibliography}

\end{document}